\documentclass[10pt,twocolumn,letterpaper]{article}

\usepackage[pagenumbers]{cvpr} 

\usepackage[dvipsnames]{xcolor}

\definecolor{cvprblue}{rgb}{0.21,0.49,0.74}
\usepackage[pagebackref,breaklinks,colorlinks,citecolor=cvprblue]{hyperref}
\usepackage{subcaption}
\usepackage{multirow}

\def\paperID{4742} 
\def\confName{CVPR}
\def\confYear{2024}

\title{One-2-3-45++: Fast Single Image to 3D Objects with Consistent Multi-View Generation and 3D Diffusion}

\author{Minghua Liu\textsuperscript{1}\thanks{Equal contribution.} \quad
Ruoxi Shi\textsuperscript{1*} \quad
Linghao Chen\textsuperscript{1,2*}\thanks{Work done during internship at UC San Diego.} \quad
Zhuoyang Zhang\textsuperscript{3*$\dag$} \quad
Chao Xu\textsuperscript{4*} \quad
Xinyue Wei\textsuperscript{1} \quad \\
Hansheng Chen\textsuperscript{5$\dag$} \quad
Chong Zeng\textsuperscript{2$\dag$} \quad
 Jiayuan Gu\textsuperscript{1} \quad
 Hao Su\textsuperscript{1} \\
\textsuperscript{1} UC San Diego \quad \textsuperscript{2} Zhejiang University \quad
\textsuperscript{3} Tsinghua University \quad
\textsuperscript{4} UCLA \quad
\textsuperscript{5} Stanford University 
}

\begin{document}

\makeatletter
\let\@oldmaketitle\@maketitle
\renewcommand{\@maketitle}{\@oldmaketitle
    \vspace{-2\baselineskip}
    \includegraphics[width=\linewidth]{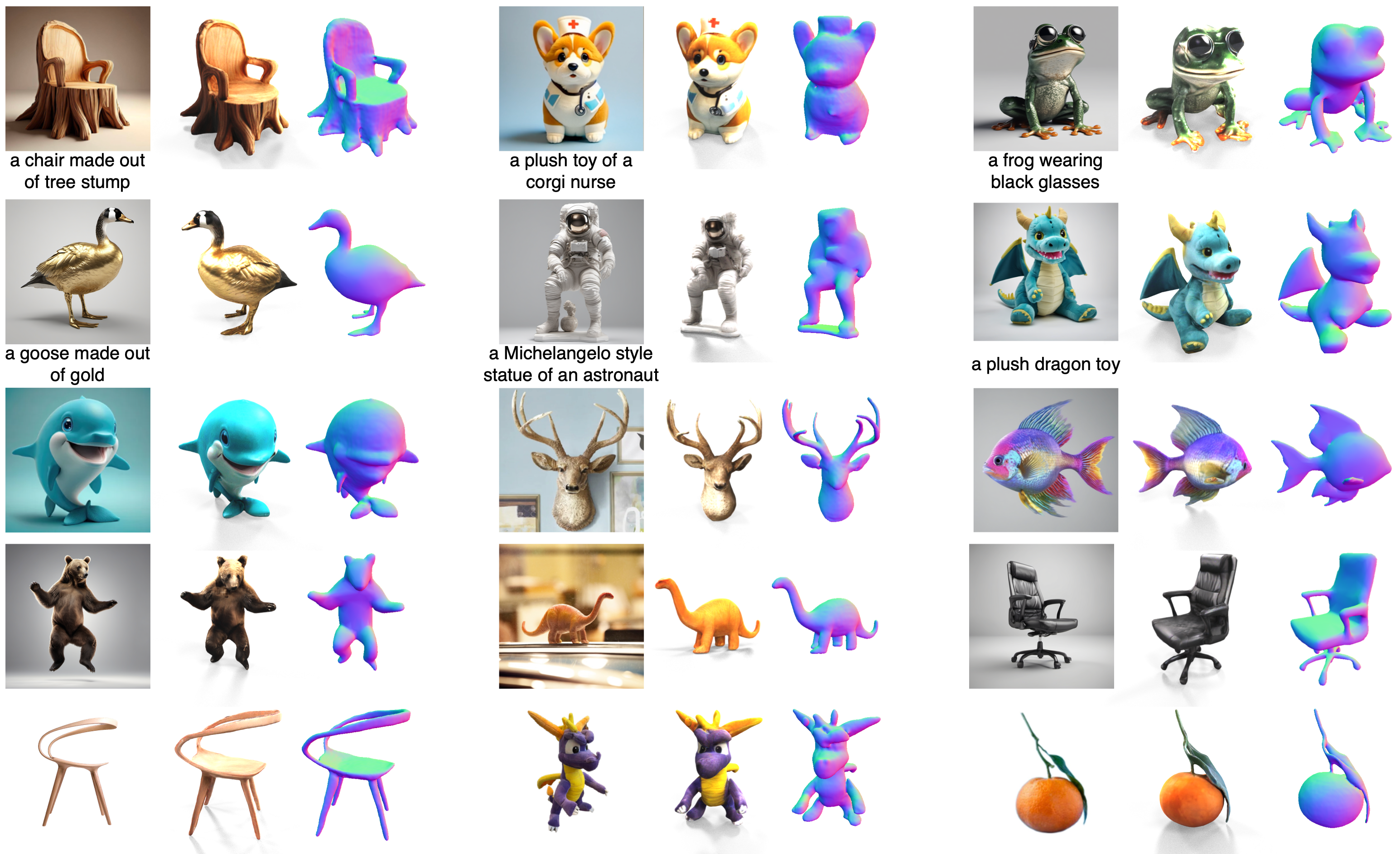}
   \vspace{-2.3em}
    \captionof{figure}{One-2-3-45++ is capable of transforming a single RGB image of any object into a high-fidelity textured mesh \textbf{in under one minute}. The generated meshes closely mirror the original input image. Input image (and text prompt), textured mesh, and normal map are shown.}
    \vspace{-1em}
    \label{fig:teaser}
\bigskip}
\makeatother

\maketitle
\urlstyle{same}

\begin{abstract}
\vspace{-1em}
Recent advancements in open-world 3D object generation have been remarkable, with image-to-3D methods offering superior fine-grained control over their text-to-3D counterparts. However, most existing models fall short in simultaneously providing rapid generation speeds and high fidelity to input images - two features essential for practical applications. In this paper, we present One-2-3-45++, an innovative method that transforms a single image into a detailed 3D textured mesh in approximately one minute. Our approach aims to fully harness the extensive knowledge embedded in 2D diffusion models and priors from valuable yet limited 3D data. This is achieved by initially finetuning a 2D diffusion model for consistent multi-view image generation, followed by elevating these images to 3D with the aid of multi-view conditioned 3D native diffusion models. Extensive experimental evaluations demonstrate that our method can produce high-quality, diverse 3D assets that closely mirror the original input image. Our project webpage: \url{https://sudo-ai-3d.github.io/One2345plus_page}.
\end{abstract}

\vspace{-2.1em}
\section{Introduction}
\label{sec:intro}
\vspace{-0.7em}

Generating 3D shapes from a single image or text prompt is a long-standing problem in computer vision and is essential for numerous applications. While remarkable progress has been achieved in the field of 2D image generation due to advanced generative methods and large-scale image-text datasets, transferring this success to the 3D domain is hindered by the limited availability of 3D training data. Although many works introduce sophisticated 3D generative models~\cite{gao2022get3d,zheng2023locally,liu2023meshdiffusion,cheng2023sdfusion}, a majority rely solely on 3D shape datasets for training. Given the limited size of publicly available 3D datasets, these methods often struggle to generalize across unseen categories in open-world scenarios.

Another line of work, exemplified by DreamFusion~\cite{poole2022dreamfusion}, Magic3D~\cite{lin2023magic3d}, harnesses the expansive knowledge or robust generative potential of 2D prior models like CLIP~\cite{radford2021learning} and Stable Diffusion~\cite{rombach2021highresolution}. They typically optimize a 3D representation (e.g., NeRF or mesh) from scratch for each input text or image. During the optimization process, the 3D representation is rendered into 2D images, and the 2D prior models are employed to calculate gradients for them. While these methods have yielded impressive outcomes, the per-shape optimization can be exceedingly time-intensive, requiring tens of minutes or even hours to generate a single 3D shape for each input. Moreover, they frequently encounter the ``multi-face'' or Janus problem, produce results with oversaturated colors and artifacts inherited from the NeRF or triplane representation, and face challenges in generating diverse results across different random seeds.


A recent work One-2-3-45~\cite{liu2023one2345} presented an innovative way to leverage rich priors of 2D diffusion models for 3D content generation. It initially predicts multi-view images via a view-conditioned 2D diffusion model, Zero123~\cite{liu2023zero}. These predicted images are subsequently processed through a generalizable NeRF method~\cite{long2022sparseneus} for 3D reconstruction. Although One-2-3-45 can produce 3D shapes in a single feed-forward pass, its efficacy is often constrained by the inconsistent multi-view predictions of Zero123, leading to compromised 3D reconstruction results. 

In this paper, we introduce One-2-3-45++, a novel method that effectively overcomes the shortcomings of One-2-3-45, delivering significantly enhanced robustness and quality. Taking a single image of any object as input, One-2-3-45++ also includes two primary stages: 2D multi-view generation and 3D reconstruction. During the initial phase, rather than employing Zero123 to predict each view in isolation, One-2-3-45++ simultaneously predicts consistent multi-view images. This is realized by tiling a concise set of six-view images into a single image and then finetuning a 2D diffusion model to generate this combined image conditioned on the input reference image. In this way, the 2D diffusion net is able to attend to each view during generation, ensuring more consistent results across views. In the second stage, One-2-3-45++ employs a multi-view conditioned 3D diffusion-based module to predict the textured mesh in a coarse-to-fine fashion. The consistent multi-view conditional images act as a blueprint for 3D reconstruction, facilitating a zero-shot hallucination capability. Concurrently, the 3D diffusion network excels in lifting the multi-view images, thanks to its ability to harness a broad spectrum of priors extracted from the 3D dataset. Ultimately, One-2-3-45++ employs a lightweight optimization technique to enhance the texture quality efficiently, leveraging consistent multi-view images for supervision.

As depicted in Fig.~\ref{fig:teaser}, One-2-3-45++ efficiently generates 3D meshes with realistic textures in under a minute, offering precise fine-grained control. Our comprehensive evaluations, including user studies and objective metrics across an extensive test set, highlight One-2-3-45++'s superiority in terms of robustness, visual quality, and, most importantly, fidelity to the input image.

\vspace{-0.5em}
\section{Related Work}
\vspace{-0.5em}
\label{sec:intro}

\begin{figure*}[t]
    \includegraphics[width=\linewidth]{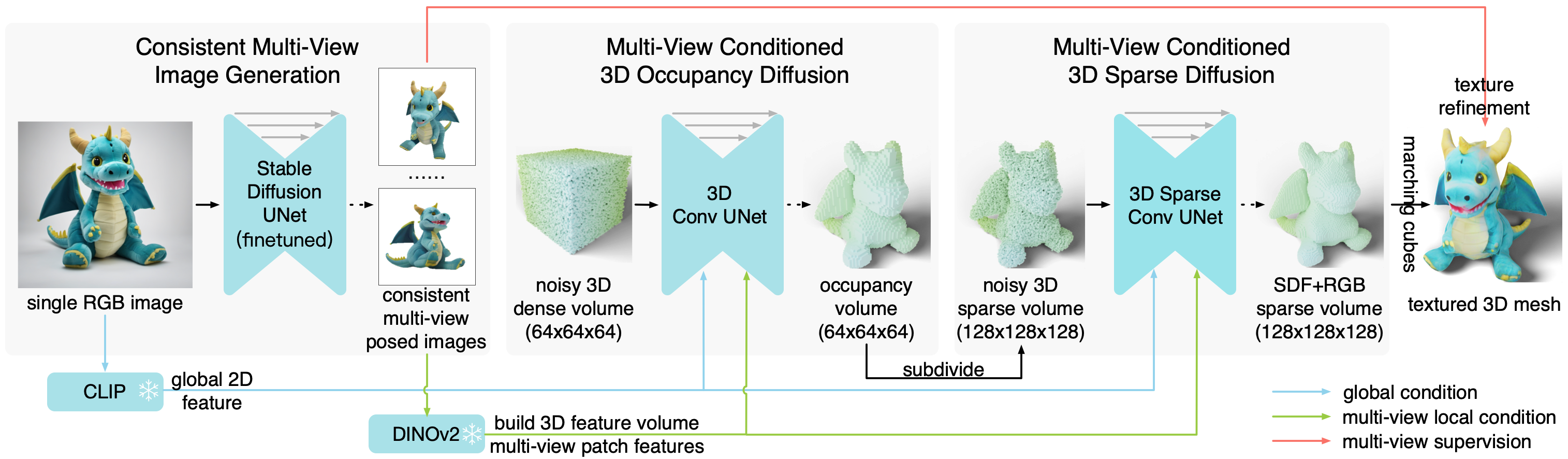}
    \vspace{-2em}
    \caption{
    Starting with a single RGB image as input, we initially produce consistent multi-view images by fine-tuning a 2D diffusion model. These multi-view images are then elevated into 3D through a pair of 3D native diffusion networks. Throughout the 3D diffusion process, the generated multi-view images act as essential guiding conditions. After extracting the 3D mesh from the denoised volume, we further enhance the texture by employing a lightweight optimization with multi-view images as supervision. Our One-2-3-45++ is capable of producing an initial textured mesh \textbf{within 20 seconds} and delivers a refined one in \textbf{roughly one minute.}}
    \vspace{-1.2em}
    \label{fig:pipeline}
\end{figure*}

\subsection{3D Generation}
\vspace{-0.5em}
3D generation has garnered significant attention in recent years. Before the advent of large-scale pre-trained 2D models, researchers often delved into 3D native generative models that learn directly from 3D synthetic data or real scans and generate various 3D representations such as point clouds~\cite{achlioptas2018learning,fan2017point,melas2023pc2,nichol2022point,zeng2022lion}, 3D voxels~\cite{choy20163d,sanghi2022clip,wu2017marrnet,xie2019pix2vox}, polygon meshes~\cite{gao2022get3d,groueix2018papier,kanazawa2018learning,nash2020polygen,liu2021deepmetahandles,wang2018pixel2mesh,liu2023meshdiffusion}, parametric models~\cite{hong2022avatarclip}, and implicit fields~\cite{jun2023shap,cheng2023sdfusion,gupta20233dgen,mescheder2019occupancy,park2019deepsdf,wu2023multiview,xu2019disn,li2023diffusion,zhang20233dshape2vecset,zheng2023locally,erkocc2023hyperdiffusion,zhao2023michelangelo,yu2023pushing}. However, given the limited availability of 3D data, these models tended to focus on a select number of categories (e.g., chairs, cars, planes, humans, etc.), struggling to generalize to unseen categories in the open world. 

The advent of recent 2D generative models (e.g., DALL-E~\cite{ramesh2021zero}, Imagen~\cite{saharia2022photorealistic}, and Stable Diffusion~\cite{rombach2022high}) and vision-language models (e.g., CLIP~\cite{radford2021learning}) has equipped us with powerful priors about our 3D world, consequently fueling a surge of research in 3D generation. Notably, models like DreamFusion~\cite{poole2022dreamfusion}, Magic3D~\cite{lin2023magic3d}, and ProlificDreamer~\cite{wang2023prolificdreamer} have pioneered a line of approach for per-shape optimization~\cite{jain2022zero,melas2023realfusion,deng2023nerdi,mohammad2022clip,lee2022understanding,metzer2023latent,michel2022text2mesh,raj2023dreambooth3d,seo2023let,tang2023make,wang2023score,xu2023neurallift,xu2023dream3d,tang2023dreamgaussian,chen2023fantasia3d,qian2023magic123,chen2023text,yu2023points}. These models are designed to optimize a 3D representation for each unique input text or image, drawing on the 2D prior models for gradient guidance. While they have yielded impressive results, these methods tend to suffer from prolonged optimization times, the "multi-face problem", oversaturated colors, and a lack of diversity in results. Some works also concentrate on creating textures or materials for input meshes, utilizing the priors of 2D models~\cite{chen2023text2tex,richardson2023texture}.

A new wave of studies, highlighted by works like Zero123~\cite{liu2023zero}, has showcased the promise of using pre-trained 2D diffusion models for synthesizing novel views from singular images or texts, opening new doors for 3D generation. For instance, One-2-3-45~\cite{liu2023one2345}, using multi-view images predicted by Zero123, can produce a textured 3D mesh in a mere 45 seconds. Nevertheless, the multi-view images produced by Zero123 lack 3D consistency. Our research, along with several concurrent studies~\cite{shi2023mvdream,liu2023syncdreamer,weng2023consistent123,lin2023consistent123,ye2023consistent,long2023wonder3d}, is dedicated to enhancing the consistency of these multi-view images – a vital step for subsequent 3D reconstruction applications.

\vspace{-0.3em}
\subsection{Sparse View Reconstruction}
\vspace{-0.5em}

While traditional 3D reconstruction methods, such as multi-view stereo or NeRF-based techniques, often demand a dense collection of input images for accurate geometry inference, many of the latest generalizable NeRF solutions~\cite{chen2021mvsnerf,johari2022geonerf,kulhanek2022viewformer,liu2022neural,long2022sparseneus,ren2023volrecon,trevithick2021grf,wang2022attention,wang2021ibrnet,yang2023contranerf} strive to learn priors across scenes. This enables them to infer NeRF from a sparse set of images and generalize to novel scenes. These methods typically ingest a few source views as input, leveraging 2D networks to extract 2D features. These pixel features are then unprojected and aggregated into 3D space, facilitating the inference of density (or SDF) and colors. However, these methods may either rely on consistent multi-view images with accurate correspondences or possess limited priors to generalize beyond training datasets.

Recently, some methods~\cite{zhou2023sparsefusion,chan2023genvs,tewari2023diffusion,karnewar2023holodiffusion} have employed diffusion models to aid sparse view reconstruction tasks. However, they generally frame the problem as novel view synthesis, necessitating additional processing, such as distillation using a 3D representation, to generate 3D content. Our work utilizes a multi-view conditioned 3D diffusion model for 3D generation. This model directly learns priors from 3D data and obviates the need for additional post-processing. Moreover, some concurrent works~\cite{long2023wonder3d,liu2023syncdreamer,shi2023mvdream} employ NeRF-based per-scene optimization for reconstruction, leveraging specialized loss functions.

\section{Method}
\vspace{-0.5em}

In traditional game studios, the creation of 3D content encompasses a series of stages, including concept art, 3D modeling, and texturing, etc. Each stage demands distinct and complementary expertise. For instance, concept artists should possess creativity, a vivid imagination, and the skill to visualize 3D assets. In contrast, 3D modelers must be skilled in 3D modeling tools and capable of interpreting and translating multi-view concept drawings into life-like models, even when drawings contain inconsistencies or errors.

One-2-3-45++ aims to harness the rich 2D priors and the valuable yet limited 3D data following a similar philosophy. As shown in Fig.~\ref{fig:pipeline}, with a single input image of an object, One-2-3-45++ starts by generating coherent multi-view images of the object. This is achieved by finetuning a pre-trained 2D diffusion model and acts akin to the role of a concept artist. These generated images are then input into a multi-view conditioned 3D diffusion model for 3D modeling. The 3D diffusion module, trained on extensive multi-view and 3D pairings, excels at converting multi-view images into 3D meshes. Finally, the produced meshes undergo a lightweight refinement module, guided by the multi-view images, to further enhance the texture quality. 

\vspace{-0.3em}
\subsection{Consistent Multi-View Generation}
\vspace{-0.5em}

Recently, Zero123 has demonstrated the potential of fine-tuning a pretrained 2D diffusion network to incorporate camera view control, thereby synthesizing novel views of an object from a single reference image. While previous studies have employed Zero123 to generate multi-view images, they often suffer from inconsistencies across different views. This inconsistency arises because Zero123 models the conditional marginal distribution for each view in isolation, without considering inter-view communication during multi-view generation. In this work, we present an innovative method to produce consistent multi-view images, significantly benefiting downstream 3D reconstruction.

\begin{figure}[t]
    \centering
    \includegraphics[width=0.9\linewidth]{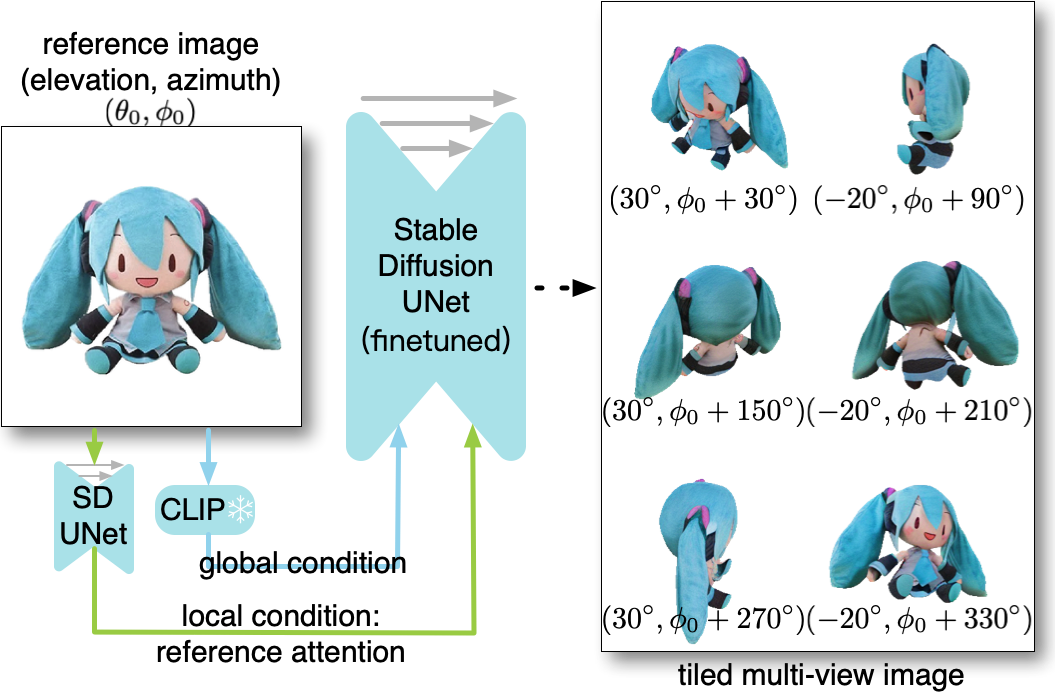}
    \vspace{-1em}
    \caption{\textbf{Consistent multi-view generation:} We stitch multi-view images into a single frame and fine-tune the Stable Diffusion model to generate this composite image, using the input reference image as conditions. We utilize predetermined absolute elevation angles and relative azimuth angles. During 3D reconstruction, we don't need to infer the elevation angle of the input image.}
    \vspace{-1.5em}
    \label{fig:multi-view-tiling}
\end{figure}

\noindent\textbf{Multi-View Tiling} To generate multiple views in a single diffusion process, we adopt a simple strategy by tiling a sparse set of 6 views into a single image with a $3 \times 2$ layout as shown in Fig.~\ref{fig:multi-view-tiling}. Subsequently, we finetune a pre-trained 2D diffusion net to generate the composite image, conditioned on a single input image. This strategy enables multiple views to interact with each other during the diffusion.

It's nontrivial to define the camera poses of the multi-view images. Given that the 3D shapes within the training dataset lack aligned canonical poses, employing absolute camera poses for the multi-view images could lead to ambiguities for the generative model. Alternatively, if we were to set the camera poses relative to the input view, as done in Zero123, downstream applications would then be required to infer the elevation angle of the input image to deduce the camera poses of the multi-view images. This additional step could introduce errors into the pipeline. To address these, we opt for fixed absolute elevation angles paired with relative azimuth angles to define the poses of multi-view images, effectively resolving the orientation ambiguity without necessitating further elevation estimation. To be more precise, the six poses are determined by alternating elevations of $30^{\circ}$ and $-20^{\circ}$, coupled with azimuths commencing at $30^{\circ}$ and incrementing by $60^{\circ}$ for each subsequent pose, as shown in Fig.~\ref{fig:multi-view-tiling}. 

\noindent\textbf{Network and Training Details} 
To fine-tune Stable Diffusion for adding image conditioning and generating coherent multi-view composite images, we employ three crucial network or training designs: (a) \textbf{Local Condition}: We adopt the reference attention technique~\cite{refattn} to incorporate local condition. Specifically, we process the reference input image with the denoising UNet model and append the self-attention key and value matrices from the conditional reference image to the corresponding attention layers of the denoising multi-view image. (b) \textbf{Global Condition}: We leverage CLIP image embedding as a global condition, replacing the text token features originally used in Stable Diffusion. These global image embeddings are multiplied by a set of learnable weights, providing the network with an overall semantic understanding of the object. (c) \textbf{Noise Schedule}: The original Stable Diffusion model was trained using a scaled-linear noise schedule. We found it necessary to switch to a linear noise scheme in our fine-tuning process.

We fine-tune the Stable Diffusion2 $v$-mode using 3D shapes from the Objaverse dataset~\cite{deitke2023objaverse}. For each shape, we generate three data points by randomly sampling the camera pose of the input image from a specified range, and selecting a random HDRI environment lighting from a curated set that offers uniform lighting. Initially, we fine-tuned only the self-attention layers along with the key and value matrices of the cross-attention layers using LoRA~\cite{hu2021lora}. Subsequently, we fine-tuned the entire UNet using a conservative learning rate. The finetuning process was conducted using 16 GPUs and took approximately 10 days.

\vspace{-0.2em}
\subsection{3D Diffusion with Multi-View Condition}
\vspace{-0.3em}

While prior work utilizes generalizable NeRF methods for 3D reconstruction, it primarily depends on accurate local correspondence of multi-view images and possesses limited priors for 3D generation. This constrains their effectiveness in lifting intricate and inconsistent multi-view images generated by the 2D diffusion network. Instead, we propose an innovative way to lift the generated multi-view images to 3D by utilizing a multi-view conditioned 3D generative model. It seeks to learn a manifold of plausible 3D shapes conditioned on multi-view images by training expressive 3D native diffusion networks on extensive 3D data.

\noindent\textbf{3D Volume Representations} As shown in Fig.~\ref{fig:pipeline}, we represent a textured 3D shape as two discrete 3D volumes, a signed distance function (SDF) volume and a color volume. The SDF volume measures the signed distance from the center of each grid cell to the nearest shape surface, while the color volume captures the color of the closest surface points relative to the center of the grid cells. Additionally, the SDF volume can be transformed into a discrete occupancy volume, where each grid cell stores a binary occupancy based on whether the absolute value of its SDF is below a predefined threshold.

\noindent\textbf{Two-Stage Diffusion} Capturing fine-grained details of 3D shapes necessitates the use of high-resolution 3D grids, which unfortunately entail substantial memory and computational costs. As a result, we follow LAS-Diffusion~\cite{zheng2023locally} to generate high-resolution volumes in a coarse-to-fine two-stage manner. Specifically, the initial stage generates a low-resolution (e.g., $n=64$) full 3D occupancy volume $F \in \mathbb{R}^{n \times n \times n \times 1}$ to approximate the shell of the 3D shape, while the second stage then generates a high-resolution (e.g., $N=128$) sparse volume $S \in \mathbb{R}^{N \times N \times N \times 4}$ which predicts fine-grained SDF values and color within the occupied area.

We employ a separate diffusion network for each stage. For the first stage, normal 3D convolution is used within the UNet to produce the full 3D occupancy volume $F$, while for the second stage, we incorporate 3D sparse convolution in the UNet to yield the 3D sparse volume $S$. Both diffusion networks are trained using the denoising loss~\cite{ho2020denoising}:

\vspace{-2em}
\begin{equation*}
 \mathcal{L}_{\boldsymbol{x}_0}=\mathbb{E}_{\boldsymbol{\epsilon} \sim \mathcal{N}(0, I), t \sim \mathcal{U}(0,1)}\left\|f\left(x_t, t, c\right)-x_0\right\|_2^2    
 \vspace{-0.5em}
\end{equation*}
where $\epsilon$ and $t$ are sampled noise and time step, $x_0$ is a data point ($F$ or $S$) and $x_t$ is its noised version, $c$ is the multi-view condition, and $f$ is the UNet. $\mathcal{N}$ and $\mathcal{U}$ denote Gaussian and uniform distribution, respectively. 

\noindent\textbf{Multi-View Condition} Training a conventional 3D native diffusion network can be challenging to generalize due to the limited availability of 3D data. However, the use of generated multi-view images can provide a comprehensive guide, greatly simplifying the imagination difficulty of 3D generation. We integrate the multi-view images to guide the diffusion process by initially extracting local image features and subsequently constructing a conditional 3D feature volume, denoted as $C$. This strategy follows the rationale that local priors facilitate easier generalization~\cite{zheng2023locally}.

\begin{figure}[t]
    \centering
    \includegraphics[width=\linewidth]{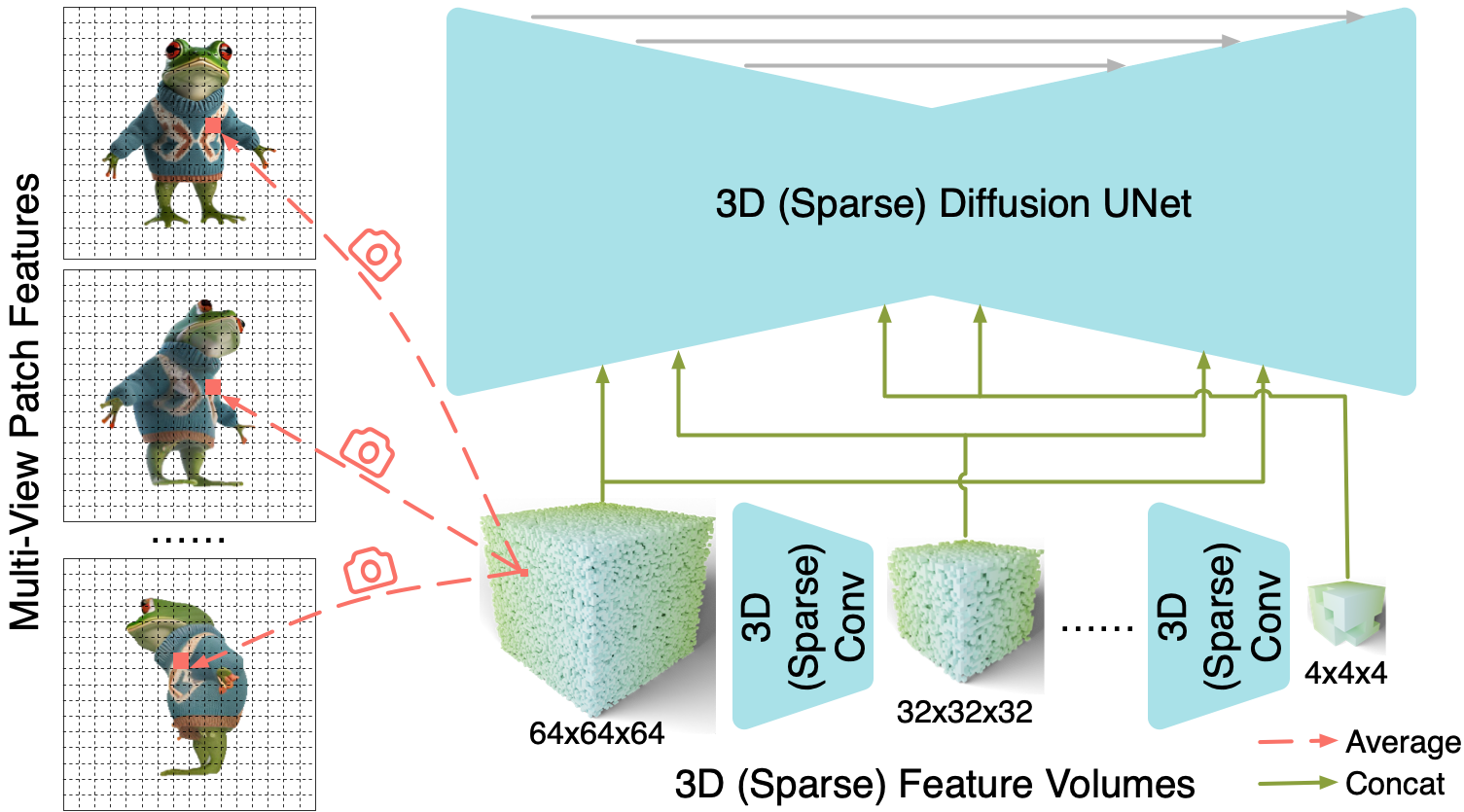}
\vspace{-2em}
 \caption{\textbf{Multi-view local condition:} We employ a pre-trained 2D backbone to extract 2D patch features for each view. These features are then aggregated using known projection matrices to construct a 3D feature volume. The volume is further processed by 3D convolutional neural networks, resulting in feature volumes of varying resolutions. Subsequently, these volumes are concatenated with the corresponding feature volumes within the diffusion U-Net to guide the 3D diffusion.}
\vspace{-1.5em}
\label{fig:multi-view condition}
\end{figure}

As shown in Fig.~\ref{fig:multi-view condition}, given $m$ multi-view images, we first employ a pretrained 2D backbone, DINOv2, to extract a set of local patch features for each image. We then build a 3D feature volume $C$ by projecting each 3D voxel within the volume onto $m$ multi-view images using the known camera poses. For each 3D voxel, we aggregate $m$ associated 2D patch features through a shared-weight MLP, followed by max pooling. These aggregated features collectively form the feature volume $C$.

In the diffusion network, the UNet consists of several levels. For example, the occupancy UNet in the initial stage has five levels: $64^3$, $32^3$, $16^3$, $8^3$, and $4^3$. Initially, we construct a conditional feature volume $C$ that matches the starting resolution, as outlined earlier. A 3D convolution network is then applied to $C$, producing volumes for the subsequent resolutions. The resultant conditional volumes are then concatenated with the volumes inside the UNet to guide the diffusion process. For the second stage, we construct sparse conditional volumes and utilize 3D sparse convolution. To benefit the diffusion of color volume, we also concatenate 2D pixel-wise projected colors to the final layer of the diffusion UNet. Moreover, we integrate the CLIP feature of the input image as a global condition. For a detailed explanation, please refer to the supplementary materials.

\noindent\textbf{Training and Inference Details} We train the two diffusion networks using 3D shapes from the Obajverse dataset~\cite{deitke2023objaverse}. For each 3D shape, we first convert it to a watertight manifold before extracting its SDF volume. We unproject the multi-view renderings of the shape to get a 3D colored point cloud, which is used to build the color volume. During training, we utilize the ground truth renderings to serve as the multi-view conditions. Since two diffusion networks are trained separately, we introduced random perturbations to camera poses and infused random noises to the initial occupancy of the second stage to enhance robustness. We train the two diffusion nets using 8xA100 GPUs for about 10 days for each stage. Please refer to the supplementary materials for more details.

During inference, a $64^3$ grid is first initialized with Gaussian noise and then denoised by the first diffusion net. Each predicted occupied voxel is further subdivided into 8 smaller voxels, used to construct a high-resolution sparse volume. The sparse volume is initialized with Gaussian noise and then denoised with the second diffusion net, resulting in predictions for the SDF and color of each voxel. The Marching Cubes algorithm is finally applied to extract a textured mesh.

\begin{figure*}[t]
    \includegraphics[width=\linewidth]{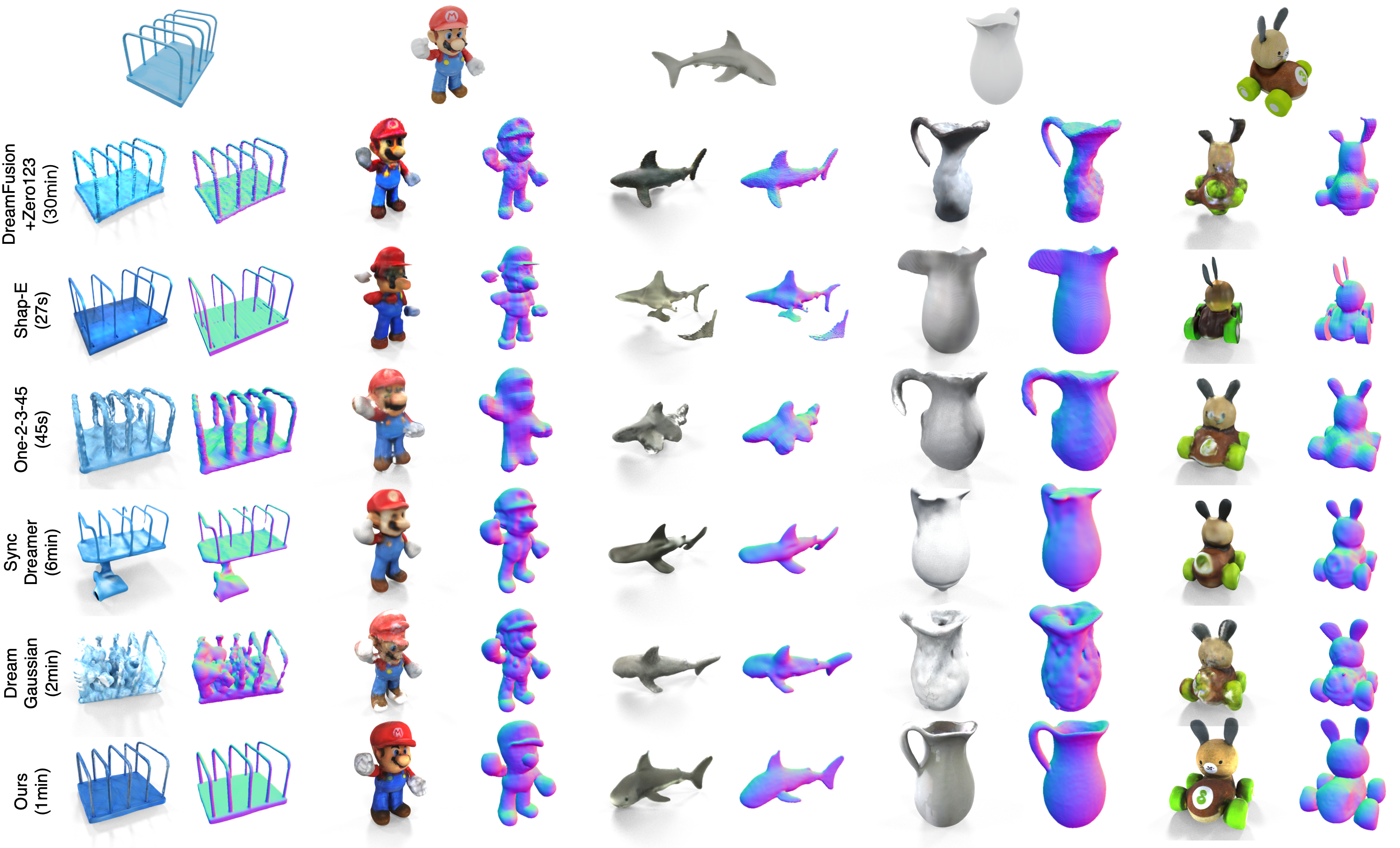}
\vspace{-2em}
    \caption{Qualitative results of various single image to 3D approaches. Input images, textured meshes, and normal maps are shown.
    }
    \label{fig:img_to_3d}
\vspace{-1em}
\end{figure*}

\vspace{-0.5em}
\subsection{Texture Refinement}
\vspace{-0.5em}

Given that multi-view images possess higher resolution than the 3D color volume, we can refine the texture of the generated mesh through a lightweight optimization process. To achieve this, we fix the geometry of the generated mesh while optimizing a color field represented by a TensoRF~\cite{chen2022tensorf}. In each iteration, the mesh is rendered to 2D by rasterization and querying the color network. We leverage the generated consistent multi-view images to guide the texture optimization using a $l2$ loss. Lastly, we bake the optimized color field onto the mesh, with the surface normal serving as the viewing direction.

\vspace{-0.5em}
\section{Experiments}
\vspace{-0.5em}

\subsection{Comparison on Image to 3D}
\vspace{-0.5em}

\noindent \textbf{Baselines:} We evaluate One-2-3-45++ against both optimization-based and feed-forward methods. Within the optimization-based approaches, our baselines include DreamFusion~\cite{poole2022dreamfusion} with Zero123 XL~\cite{liu2023zero} as its backbone, as well as SyncDreamer~\cite{liu2023syncdreamer}, and DreamGaussian~\cite{tang2023dreamgaussian}. For feed-forward approaches, we compare with One-2-3-45~\cite{liu2023one2345} and Shap-E~\cite{jun2023shap}. We employ the ThreeStudio~\cite{threestudio2023} implementation for Zero123 XL~\cite{threestudio2023} and the original official implementations for the other methods.
\begin{table}[t]
\setlength{\tabcolsep}{2pt}
\footnotesize
  \centering
  \caption{Comparison on single image to 3D. Evaluated on the GSO~\cite{downs2022google} dataset, which contains 1,030 3D objects.}
  \vspace{-1em}
    \begin{tabular}{c|cccc}
    \toprule
    Method & F-Sco. (\%)$\uparrow$ & CLIP-Sim$\uparrow$ & User-Pref. (\%)$\uparrow$ & Time$\downarrow$ \\
    \midrule
    Zero123 XL~\cite{deitke2023objaversexl} &  91.6    &   73.1    &   58.6    &  30min\\
    One-2-3-45~\cite{liu2023one2345} &   90.4    & 70.8      &    52.7   & 45s \\
    SyncDreamer~\cite{liu2023syncdreamer} & 84.8  &    68.9   &  28.4     & 6min \\
    DreamGaussian~\cite{tang2023dreamgaussian} & 81.0    & 68.4  &    31.5   & 2min \\
    Shap-E~\cite{jun2023shap} & 91.8  & 73.1  &    40.8   & 27s \\
    Ours  & \textbf{93.6} & \textbf{81.0} &    \textbf{87.6}   & 60s \\
    \bottomrule
    \end{tabular}%
    \vspace{-1.5em}
  \label{tab:img_com}%
\end{table}%

\noindent \textbf{Dataset and Metrics:} We assess the performance of the methods using the entire set of 1,030 shapes from the GSO dataset \cite{downs2022google}, which were not exposed to any of the methods during training to the best of our knowledge.  For each shape, we generate a frontal view image to serve as the input. In line with One-2-3-45 \cite{liu2023one2345}, we employ the F-Score and CLIP similarity as our evaluation metrics. The F-Score evaluates the geometric similarity between the predicted mesh and the ground truth mesh. For the CLIP similarity metric, we render 24 different views for each predicted and ground truth mesh, compute the CLIP similarity for each corresponding pair of images, and then average these values across all views. Prior to metric computation, we align the predicted mesh with the ground truth mesh using a combination of linear search and the ICP algorithm. 

\begin{figure}[t]
    \includegraphics[width=\linewidth]{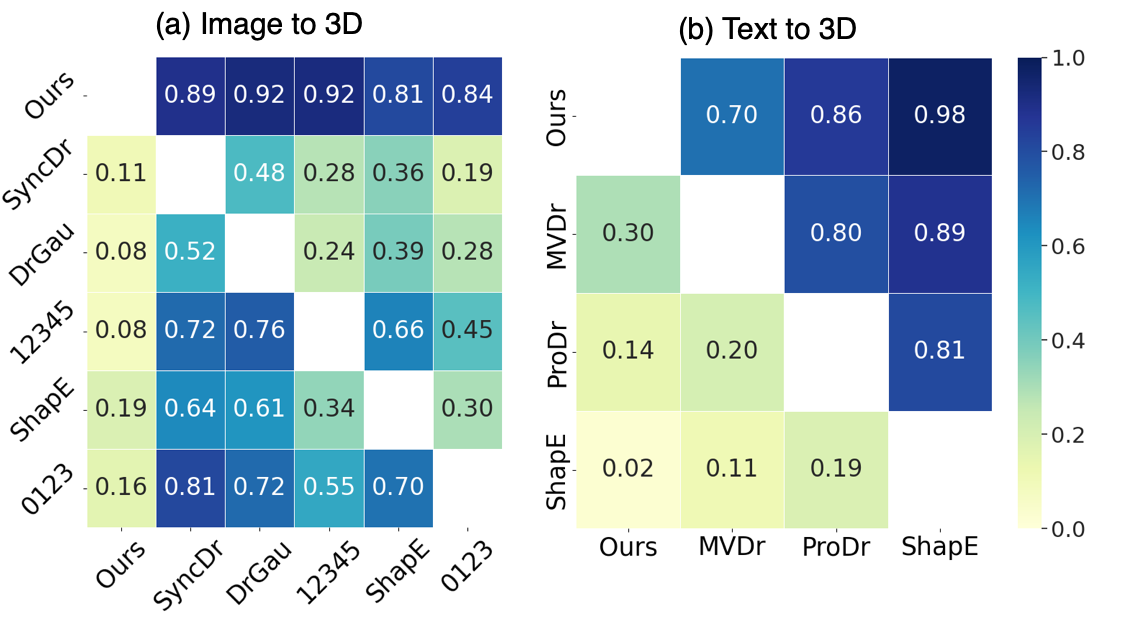}
    \vspace{-2em}
    \caption{Results of a user study involving 53 participants. Each cell displays the probability or preference rate at which one method (row) outperforms another (column).
    }
    \vspace{-1.5em}
    \label{fig:user_study}
\end{figure}

\begin{figure*}[t]
    \includegraphics[width=\linewidth]{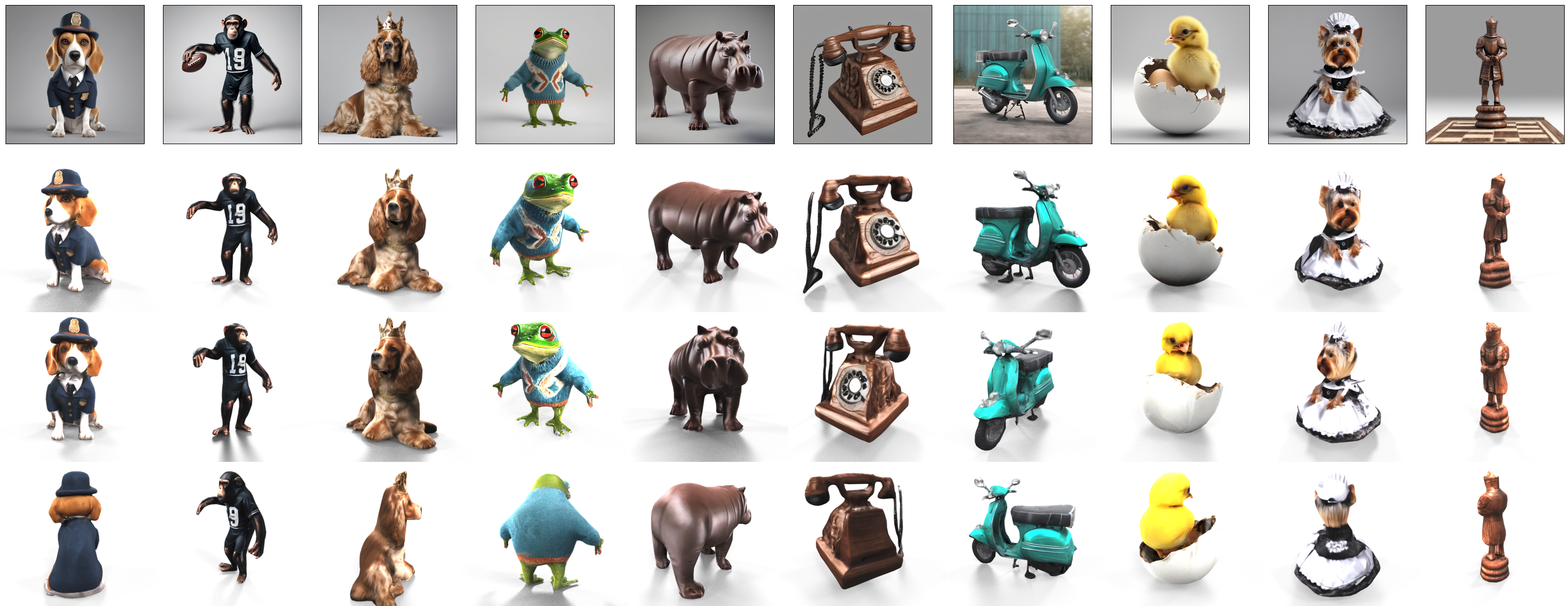}
    \vspace{-2em}
    \caption{Our qualitative results: top row displays input images; subsequent rows shows multi-view renderings of the generated meshes. 
    }
    \label{fig:qualitative}
    \vspace{-1em}
\end{figure*}

\noindent \textbf{User Study:} A user study was also carried out. For each participant, 45 shapes were randomly selected from the entire GSO dataset, and two methods were randomly sampled for each shape. Participants were asked to choose the result from each pair of comparative outcomes that exhibits superior quality and better aligns with the input image. The preference rate for all methods was then tallied based on these selections. In total, 2,385 evaluated pairs were collected from 53 participants.

\noindent \textbf{Results:} As presented in Tab.~\ref{tab:img_com}, One-2-3-45++ surpasses all baseline methods regarding F-Score and CLIP similarity. The user preference scores further highlight a significant performance disparity, with our method outperforming competing approaches by a substantial margin. Refer to Fig.~\ref{fig:user_study} for an in-depth confusion matrix, which illustrates that One-2-3-45++ outperforms One-2-3-45 92\% of the time. Moreover, when compared to optimization-based methods, our approach demonstrates notable runtime advantages. Fig.~\ref{fig:img_to_3d} and~\ref{fig:qualitative} show qualitative results.

\vspace{-0.5em}
\subsection{Comparison on Text to 3D}
\vspace{-0.5em}

\begin{figure*}[t]
    \includegraphics[width=\linewidth]{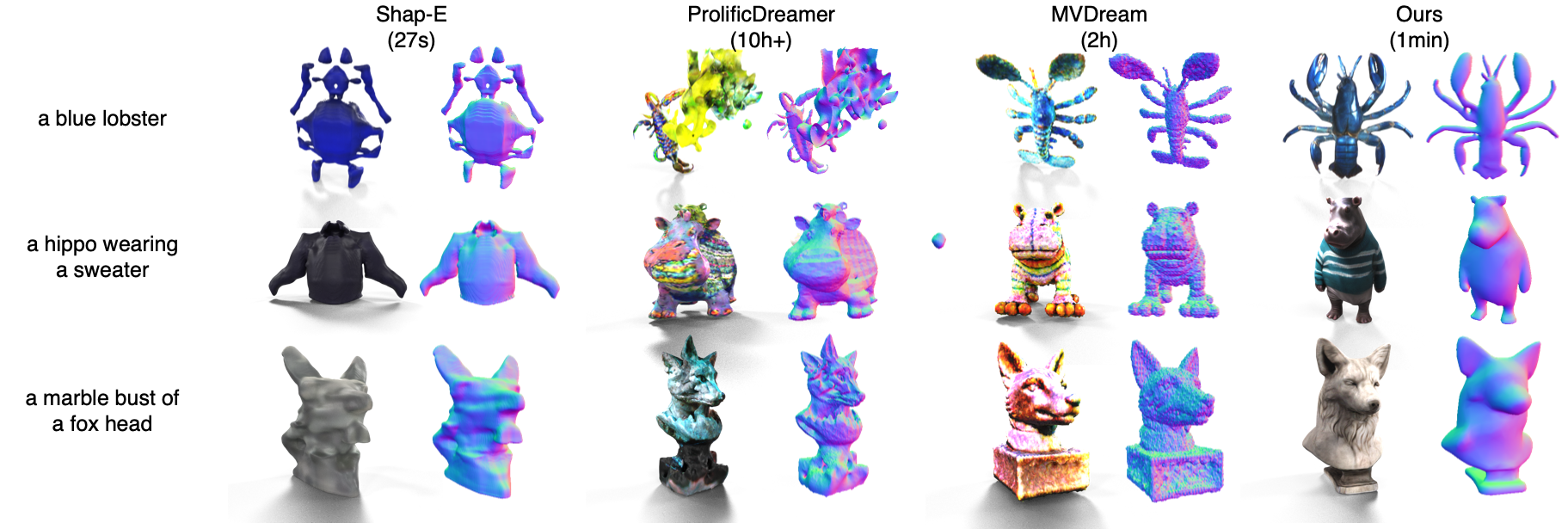}
    \vspace{-2em}
    \caption{
    Qualitative results of various text to 3D approaches. Input images, textured meshes, and normal maps are shown.
    }
    \label{fig:text_to_3d}
    \vspace{-1.5em}
\end{figure*}

\noindent \textbf{Baselines:} We compared One-2-3-45++ with optimization-based methods, specifically ProlificDreamer~\cite{wang2023prolificdreamer} and MVDream~\cite{shi2023mvdream}, as well as a feed-forward approach, Shap-E~\cite{jun2023shap}. For ProlificDreamer, we utilized the ThreeStudio implementation~\cite{threestudio2023}, while for the remaining methods, we employed their respective official implementations.

\begin{table}[t]
\setlength{\tabcolsep}{3pt}
\footnotesize
  \centering
  \caption{Quantitative comparison with various text to 3D methods. Evaluated on 50 text prompts from DreamFusion~\cite{poole2022dreamfusion}.}
  \vspace{-1em}
    \begin{tabular}{c|ccc}
    \toprule
    Method & CLIP-Sim$\uparrow$ & User-Pref.$\uparrow$ & Runtime$\downarrow$ \\
    \midrule
    ProlificDreamer~\cite{wang2023prolificdreamer} &  25.7     &   39.5    & 10h+ \\
    MVDream~\cite{shi2023mvdream} &   24.8    &  66.2     & 2h \\
    Shap-E~\cite{jun2023shap} &  22.3     &   11.1    & 27s \\
    Ours  &  \textbf{26.8}     &  \textbf{84.1}     & ~60s \\
    \bottomrule
    \end{tabular}%
    \vspace{-1.5em}
  \label{tab:text_com}%
\end{table}%

\noindent \textbf{Dataset and Metrics:} Given that many baseline approaches necessitate hours to produce a single 3D shape, our evaluation was conducted on 50 text prompts, sampled from DreamFusion~\cite{poole2022dreamfusion}. We utilize CLIP similarity, calculated by comparing 24 rendered views of the predicted mesh against the input text prompt and then averaging the similarity scores across all views.

\noindent \textbf{User Study:} The user study, akin to the image-to-3D evaluation, involved 30 pairs of outcomes randomly selected for each participant. In total, 1,590 evaluation pairs were collected from 53 participants.

\noindent \textbf{Results:} As illustrated in Tab.~\ref{tab:text_com}, One-2-3-45++ outperforms all baseline methods in terms of CLIP similarity. This is further corroborated by user preference scores, with our method significantly outshining rival techniques. See Fig.~\ref{fig:user_study} for an in-depth analysis. When directly comparing One-2-3-45++ with the second-best method, MVDream~\cite{shi2023mvdream}, our approach commands a 70\% user preference rate. Moreover, while our method delivers prompt results, MVDream~\cite{shi2023mvdream} requires about 2 hours to generate a single shape. Fig.~\ref{fig:text_to_3d} shows qualitative results.

\vspace{-0.5em}
\subsection{Analyses}
\vspace{-0.5em}

\noindent \textbf{Ablation Studies of Overall Pipeline} One-2-3-45++ is comprised of three key modules: consistent multi-view generation, multi-view conditioned 3D diffusion, and texture refinement. We conducted ablation studies on these modules using the complete GSO dataset~\cite{downs2022google}, with results detailed in Tab.~\ref{tab:ablation}. Replacing our consistent multi-view generation module with Zero123XL~\cite{deitke2023objaversexl} led to a noticeable performance decline. Furthermore, substituting our 3D diffusion module with the generalizable NeRF used in One-2-3-45~\cite{liu2023one2345} resulted in an even more significant performance drop. However, the inclusion of our texture refinement module markedly improved texture quality, yielding higher CLIP similarity scores.

\begin{table}[t]
\setlength{\tabcolsep}{1.5pt}
\footnotesize
  \centering
  \caption{Ablation studies of different modules. Evaluated on the complete GSO~\cite{downs2022google} dataset. ``MultiView'', ``Reconstruction'', and ``Texture'' indicate multi-view generation, sparse view reconstruction, and texture refinement modules, respectively.}
    \vspace{-0.5em}
    \begin{tabular}{ccc|ccc}
    \toprule
    MultiView & Reconstruction & Texture & F-Sc.$\uparrow$ & CLIP-Sim$\uparrow$ & Time$\downarrow$ \\
    \midrule
    Zero123 XL~\cite{deitke2023objaversexl} & Ours     & w/o     &  92.9     &  71.9     & 14s \\
    Ours     & SparseNeuS~\cite{long2022sparseneus} & w/o     & 81.2      &   67.2    & 15s \\
    Ours     & Ours     & w/o     &   \textbf{93.6}    &   73.4    & 20s \\
    Ours     & Ours     & w/     & \textbf{93.6} & \textbf{81.0} & 60s \\
    \bottomrule
    \end{tabular}%
    \vspace{-1.5em}
  \label{tab:ablation}%
\end{table}%

\noindent \textbf{Ablation Studies of 3D Diffusion} Tab.~\ref{tab:diffusion_ablation} presents the results of an ablation study of the 3D diffusion module. The study highlights the importance of multi-view images for the module's efficacy. When the module operates without multi-view conditions, relying solely on the global CLIP feature from a single input view (rows a and f), there is a significant decline in performance. Conversely, the One-2-3-45++ approach leverages multi-view local features by constructing a 3D feature volume with known projection matrices. A mere concatenation of global CLIP features from multiple views also impairs performance (rows b and f), underlining the value of multi-view local conditions. Global CLIP features of the input view, however, provide global shape semantics; their removal results in decreased performance (rows c and e). Although One-2-3-45++ uses predicted multi-view images for 3D reconstruction, incorporating these predicted images during training of the 3D diffusion module can lead to a performance downturn (rows d and e) due to the potential mismatch between the predicted multi-view images and actual 3D ground truth meshes. To train the module effectively, we utilize ground truth renderings. Recognizing that predicted multi-view images may be flawed, we introduce random perturbations to projection matrices during training to enhance robustness when processing predicted multi-view images (rows e and f).

\definecolor{iglGreen}{RGB}{153,203,67}
\begin{table}[t]
\setlength{\tabcolsep}{2pt}
\footnotesize
  \centering
  \caption{Ablation study of the 3D diffusion module. 3D IoU of the initial-stage occupancy prediction is reported. Note that the 3D IoU is computed for the 3D shell, excluding the solid interior.}
  \vspace{-1em}
    \begin{tabular}{c|cccc|c}
    \toprule
   id & multi-view cond. & global cond. & image source & proj. perturb. & 3D IoU $\uparrow$ \\
    \midrule
   a &  w/o   & w/    & rendering & N/A   & 18.3 \\
   b &  global & w/    & rendering & N/A   & 24.4 \\
   c & local & w/o   & rendering & w/o    & 41.4 \\
   d & local & w/    & prediction & w/o    & 41.9 \\
   e & local & w/    & rendering & w/o   & 44.1  \\
   f & local & w/    & rendering & w/    & 45.1 \\
    \bottomrule
    \end{tabular}%
\vspace{-1.5em}
  \label{tab:diffusion_ablation}%
\end{table}%

\noindent \textbf{Comparison on Multi-View Generation} We also evaluate our consistent multi-view generation module against existing approaches, namely Zero123~\cite{liu2023zero} and its scaled variant~\cite{deitke2023objaversexl}, alongside two concurrent works: SyncDreamer~\cite{liu2023syncdreamer} and Wonder3D~\cite{long2023wonder3d}. Our comparison utilizes the GSO~\cite{downs2022google} dataset, where for each object, we render a single input image and task the methods with producing multi-view images. For Zero123 and Zero123 XL, we utilize the same target poses as our approach. However, for Wonder3D and SyncDreamer, we employ the target poses preset by these methods, as they do not support altering camera positions during inference. As presented in Tab.~\ref{tab:multi-view-com}, our approach surpasses current methodologies in PSNR, LPIPS, and foreground mask IoU. Notably, Wonder3D~\cite{long2023wonder3d} employs orthographic projection in its training phase, which compromises its robustness when dealing with perspective images during inference. SyncDreamer~\cite{liu2023syncdreamer} only generates views at an elevation of $30^{\circ}$, a simpler setting than ours. Moreover, since these metrics do not assess 3D consistency across views, please refer to supplementary for additional qualitative comparisons and discussions.

\begin{table}[t]
\footnotesize
  \centering
  \setlength{\tabcolsep}{2.5pt}
  \caption{Comparison of different multi-view generation methods. Evaluated on the complete GSO~\cite{downs2022google} dataset.}
  \vspace{-1em}
    \begin{tabular}{c|c|ccc}
    \toprule
   &   Target Elevations    & PSNR $\uparrow$ & LPIPS $\downarrow$ & Mask IoU $\uparrow$ \\
    \midrule
    Zero123~\cite{liu2023zero} & \multirow{3}[0]{*}{$30^{\circ}$ and $-20^{\circ}$} &  20.32    &  \textbf{0.110} & 0.856  \\ 
    Zero123 XL~\cite{deitke2023objaversexl} & &20.11  & 0.113   & 0.869    \\ 
    Ours  &  & \textbf{22.12}     & \textbf{0.110}   & \textbf{0.878}  \\ 
    \hline
    SyncDreamer~\cite{liu2023syncdreamer} & $30^{\circ}$ & 21.67  & 0.095   &   0.894  \\ 
    \hline
    Wonder3D~\cite{long2023wonder3d} & $0^{\circ}$ & 18.67  & 0.130  & 0.635 \\      
    \bottomrule
    \end{tabular}%
    \vspace{-1.5em}
  \label{tab:multi-view-com}%
\end{table}%

\vspace{-0.5em}
\section{Conclusion}
\vspace{-0.5em}
In this paper, we introduced One-2-3-45++, an innovative approach for transforming a single image of any object into a 3D textured mesh. This method stands out by offering more precise control compared to existing text-to-3D models, and it is capable of delivering high-quality meshes swiftly—typically in under 60 seconds. Additionally, the generated meshes exhibit a high fidelity to the original input image. Looking ahead, there is potential to enhance the robustness and detail of the geometry by incorporating additional guiding conditions from 2D diffusion models, alongside RGB images.

\newpage

\section*{Acknowledgments}
We would like to thank Google Cloud and Lambda Labs for their invaluable support in providing computing resources.

{
    \small
    \bibliographystyle{ieeenat_fullname}
    \bibliography{main}
}


\end{document}